# A Deep Learning Framework for Sequence Mining with Bidirectional LSTM and Multi-Scale Attention

Tao Yang
Illinois Institute of Technology
Chicago, USA

Yu Cheng
Fordham University
New York, USA

Yaokun Ren
Northeastern University
Seattle, USA

Yujia Lou
University of Rochester
Rochester, USA

Minggu Wei
University of Saskatchewan
Saskatoon, Canada

Honghui Xin*
Northeastern University
Seattle, USA

*Abstract-This paper addresses the challenges of mining latent patterns and modeling contextual dependencies in complex sequence data. A sequence pattern mining algorithm is proposed by integrating Bidirectional Long Short-Term Memory (BiLSTM) with a multi-scale attention mechanism. The BiLSTM captures both forward and backward dependencies in sequences, enhancing the model's ability to perceive global contextual structures. At the same time, the multi-scale attention module assigns adaptive weights to key feature regions under different window sizes. This improves the model's responsiveness to both local and global important information. Extensive experiments are conducted on a publicly available multivariate time series dataset. The proposed model is compared with several mainstream sequence modeling methods. Results show that it outperforms existing models in terms of accuracy, precision, and recall. This confirms the effectiveness and robustness of the proposed architecture in complex pattern recognition tasks. Further ablation studies and sensitivity analyses are carried out to investigate the effects of attention scale and input sequence length on model performance. These results provide empirical support for structural optimization of the model.*

## I. INTRODUCTION

In the era of information explosion, data is growing at an exponential rate. Sequence data, such as financial transactions, user behavior traces, sensor readings, medical monitoring records, and natural language texts, are especially prominent. Their widespread distribution and temporal nature have made sequence data mining a vital direction in the field of data mining. Sequential pattern mining aims to automatically discover hidden temporal dependencies and behavioral patterns from ordered data streams [1]. This is crucial for enabling data-driven predictive analysis, behavior modeling, and intelligent decision support. However, traditional approaches often rely on heuristic rules or shallow statistical features. These methods struggle with high-dimensional, nonlinear, and multi-scale characteristics in complex sequence structures, which limits their performance and generalization in real-world tasks.

In recent years, deep learning techniques have advanced rapidly, especially in natural language processing and time series modeling. Models based on Recurrent Neural Networks (RNNs) have shown strong capabilities in sequence mining tasks. Among them, Bidirectional Long Short-Term Memory (BiLSTM), as an extension of the LSTM architecture, captures both forward and backward dependencies in sequences [2]. This significantly improves the perception of global contextual information and suits complex sequence data with long-range dependencies. However, BiLSTM still faces limitations in multi-scale information modeling. In scenarios where patterns of different granularities coexist and local patterns play key roles, the model's ability to focus on salient regions remains insufficient [3].

To enhance the representational and attention capabilities of sequence mining algorithms, attention mechanisms have been gradually integrated into sequence modeling frameworks. Attention mechanisms dynamically adjust the model's focus across different positions in the input sequence. They enable the model to concentrate on key patterns from various dimensions or granularities, allowing dynamic allocation of feature importance and differentiated modeling of contextual information. In multi-scale attention mechanisms, attention modules at different scales model the sequence from both local and global perspectives. This helps extract more discriminative and expressive features while preserving the integrity of original information. Such mechanisms provide deep neural networks with flexible structures, making them more robust and adaptive to complex structured sequences [4].

The integration of BiLSTM and multi-scale attention mechanisms offers a novel approach for sequence pattern mining. BiLSTM effectively models bidirectional dependencies in temporal data [5], while multi-scale attention captures important patterns at various levels. Their combination not only enhances the model's expressiveness but also strengthens its ability to identify salient features in complex sequences [6]. Moreover, this hybrid structure is highly transferable and generalizable. It can be widely applied to tasks such as text classification, fault prediction, and user behavior recognition, further expanding the application scope of deep learning models in data mining.

Given the above context, exploring a sequence pattern mining algorithm based on BiLSTM and multi-scale attention mechanisms holds significant theoretical and practical value. On one hand, this research provides a more refined and interpretable deep mining method for complex sequence modeling, extending the reach of existing pattern recognition techniques [7]. On the other hand, it offers a reliable algorithmic foundation for intelligent systems to process temporal information, perform dynamic prediction, and support behavior modeling. As artificial intelligence and data mining continue to converge, proposing efficient and robust models for structurally complex and semantically rich sequence data has become a critical step toward advancing intelligent applications.

## II. METHOD

The algorithm architecture proposed in this study is mainly composed of a bidirectional LSTM layer and a multi-scale attention mechanism. Its model architecture is shown in Figure 1.

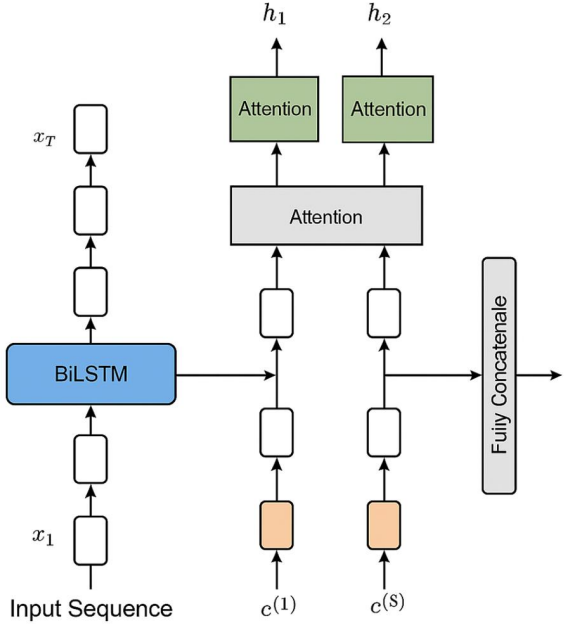

Figure 1. Overall model architecture

Figure 1 shows the overall structure of the proposed sequence pattern mining model that integrates bidirectional LSTM and multi-scale attention mechanisms. The model first receives the input sequence and extracts bidirectional temporal features through the BiLSTM layer to fully capture the dynamic dependencies between contexts. Subsequently, the extracted features are input into the attention modules of multiple scales [8]. The attention mechanisms of each scale perform weighted modeling on the important local or global information in the sequence according to different receptive fields to generate a set of context vectors $c^{(1)}, c^{(2)}, ..., c^{(S)}$. These vectors, after being concatenated and integrated, are passed through a fully connected layer to complete the recognition and mining of sequence patterns. This architecture enables a synergistic fusion of temporal modeling and multi-scale feature perception, which is critical for identifying salient patterns embedded within complex sequential data. The use of a BiLSTM structure ensures the model captures temporal dependencies in both directions, while the adaptive weighting strategy—guided by a multi-scale attention mechanism—facilitates the selective emphasis of features across varying temporal resolutions. This design draws on principles from recent advancements in sparse decomposition and adaptive fusion for multimodal data mining [9], which highlight the importance of dynamically weighting heterogeneous feature representations to enhance interpretability and performance. Moreover, the framework reflects contemporary directions in distributed and privacy-preserving modeling [10], where hierarchical representations and joint optimization strategies are employed to preserve structural richness while improving predictive robustness. The representation formulation and attention weighting also benefit from data-driven factor modeling techniques applied in sequence forecasting tasks [11], reinforcing the structural coherence and adaptability of the proposed approach.

First, let the input sequence be $X = \{x_1, x_2, ..., x_T\}$, where $x_t \in R^d$ represents the feature vector at time step t, and the sequence length is T. Through the BiLSTM structure, the model can capture the forward and backward time-dependent information respectively [12], with the forward hidden state being $\vec{h}_t$, the backward hidden state being $\overleftarrow{h}_t$, and the final hidden representation being $h_t = [\vec{h}_t, \overleftarrow{h}_t] \in R^{2h}$. This structure enhances the ability to model context and helps learn global semantic associations. The hidden state sequence $H = \{h_1, h_2, ..., h_T\}$ output by this layer will serve as the input of the subsequent attention mechanism.

In order to further improve the model's ability to perceive key time sequence segments, a multi-scale attention mechanism is introduced [13]. Specifically, there are S attention heads of different scales, each scale corresponds to a specific window width $w_s$, which is used to extract local or global sequence context features. For the sth scale, its attention weight is calculated by the following formula:

$$a_t^{(s)} = \frac{\exp(e_t^{(s)})}{\sum_{k=t-w_s}^{t+w_s} \exp(e_k^{(s)})}, e_t^{(s)} = \tanh(W_s h_t + b_s)$$

Where $W_s \in R^{1 \times 2h}, b_s \in R$ is a trainable parameter and $a_t^{(s)}$ represents the attention score at the sth scale at the current time step. Next, a weighted context representation is extracted for each scale:

$$c(s) = \sum_{t=1}^{T} a_t^{(s)} h_t$$

Finally, the context representations of all scales are concatenated to obtain a unified multi-scale feature representation $c = [c^{(1)}; c^{(2)}, ..., c^{(S)}]$.

Based on the above fusion structure, the model realizes the joint mining of time-dependent modeling and semantic information at different scales. In downstream tasks, the final feature representation c will be input into the fully connected layer for classification or regression. The loss function depends on the specific task. If it is a classification task, the cross entropy loss function is used:

$$L = -\sum_{i=1}^{C} y_i \log(y'_i), \quad y' = \text{softmax}(W_c c + b_c)$$

Where C represents the number of categories, $y_i$ is the one-hot encoding of the true label, $y'$ is the model prediction probability, and $W_c$、$b_c$ is the fully connected layer parameter. Through end-to-end training, the model not only captures sequence features but also dynamically adapts to information needs of different scales, improving the ability to mine complex sequence patterns and generalization performance.

### III. EXPERIMENT

#### A. Datasets

This study uses the Learning Gesture dataset from the UEA Multivariate Time Series Classification Archive as the experimental foundation. Released by the University of East Anglia, this dataset is widely used in sequence modeling and pattern recognition tasks [14]. It is both representative and challenging. The data is collected from a motion capture system to describe dynamic hand gestures performed by individuals over time. It exhibits strong temporal dependencies, making it suitable for evaluating sequence pattern mining models.

The Learning Gesture dataset contains 20 gesture categories. Each sample sequence consists of synchronized data from multiple sensors. The dimensionality is 5, corresponding to five sensor channels. The average sequence length ranges from 100 to 200 time steps. The data shows high variability and redundancy. It includes strong local patterns and contextual dependencies, which reflect the characteristics of complex temporal data in real-world environments. The original dataset is pre-divided into training and testing sets, which facilitates model training and generalization evaluation.

By using this dataset, the proposed algorithm is validated in terms of its effectiveness and robustness on real-world multivariate time series data. The results also demonstrate the adaptability of the BiLSTM combined with multi-scale attention mechanisms in multi-class pattern mining tasks. The fine-grained differences between gesture categories impose strict demands on the model's ability to capture details and model temporal structures. This provides a solid basis for evaluating algorithm performance in complex real-world scenarios.

#### B. Experimental Results

This paper first gives an accuracy comparison experiment of different models in multi-category sequence recognition tasks. The experimental results are shown in Table 1.

Table 1. Comparison of the accuracy of different models in multi-category sequence recognition tasks

| Model | Acc | Precision | Recall |
|---|---|---|---|
| BiLSTM + Multi-Scale Attention(Ours) | 94.27 | 93.88 | 94.10 |
| Informer[15] | 91.52 | 91.20 | 90.85 |
| TimesNet[16] | 92.76 | 92.10 | 91.87 |
| FEDformer[17] | 89.41 | 89.90 | 88.75 |
| TSMixer [118] | 90.83 | 90.33 | 90.05 |

As shown in the experimental results in Table 1, the proposed model, which integrates Bidirectional LSTM and a multi-scale attention mechanism, achieves the best performance in multi-class sequence recognition tasks. It outperforms all baseline models in terms of accuracy, precision, and recall. The accuracy reaches 94.27%, significantly higher than recently proposed models such as Informer and TimesNet. This demonstrates the model's stronger capability in capturing complex sequence patterns. The results indicate that combining bidirectional temporal modeling with a multi-scale attention mechanism can more effectively extract key patterns and enhance the perception of important sequence information.

In comparison, models based on Transformer variants, such as Informer and FEDformer, possess strong modeling capacities. However, when dealing with short to medium-length sequences with strong temporal dependencies, their unidirectional attention structure shows limitations in feature focusing. This leads to weaker recall performance compared to the proposed model. Moreover, although TimesNet and TSMixer introduce novel structural optimizations for time series modeling, they still fall short in multi-class recognition tasks. This may result from an imbalance in capturing both global and local information, making it difficult to recognize fine-grained class features.

Overall, the proposed model achieves performance improvements across different evaluation metrics. This fully validates the effectiveness of the multi-scale attention mechanism in enhancing the temporal modeling capacity of BiLSTM. The experimental results not only demonstrate the superiority of this architecture in sequence mining tasks but also confirm its good generalization ability and stability. It is well-suited for broader application in other complex and diverse time series recognition scenarios.

Next, an experiment is conducted to investigate the effect of different sequence lengths on the robustness of the model. The experimental results are shown in Figure 2.

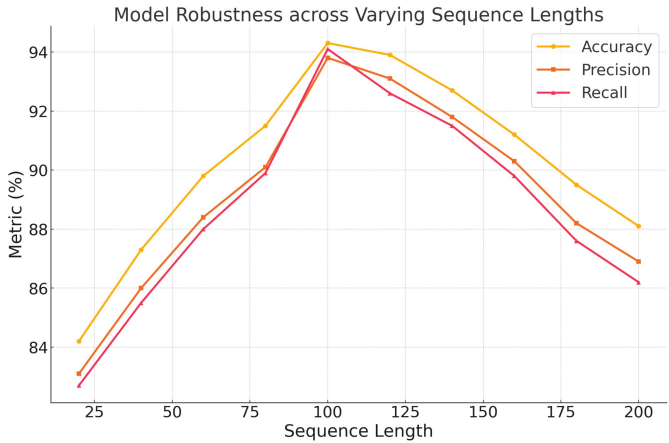

Figure 2. Model Robustness across Varying Sequence Lengths

As shown in the experimental results in Figure 2, the model exhibits notable differences in robustness under varying sequence lengths. As the sequence length increases, accuracy, precision, and recall first improve and then decline. This indicates that sequence length affects both the completeness and redundancy of feature learning. The model achieves optimal performance at a sequence length of 100, suggesting that this range provides the most suitable information density and temporal dependencies for the proposed architecture.

In the short sequence range (20 to 60), the model performs relatively poorly. This is mainly because short sequences fail to capture complete contextual structures. As a result, BiLSTM cannot effectively model long-term dependencies, and the attention mechanism struggles to focus on meaningful features. When the sequence is too long (e.g., 180 to 200), it contains more information but also introduces noise and redundant features. This reduces training efficiency and causes information dilution during weight assignment, leading to a drop in performance metrics.

Overall, the results confirm that the proposed model demonstrates strong expressiveness and robustness with medium-length sequences. It effectively captures key patterns and maintains stable performance. These findings also suggest that, in practical applications, sequences should be appropriately trimmed or padded according to the task context. This helps match the model's optimal recognition window and improves overall mining effectiveness and system performance.

Finally, a sensitivity analysis of the effect of window size on the results in the multi-scale attention module is given, and the experimental results are shown in Figure 3.

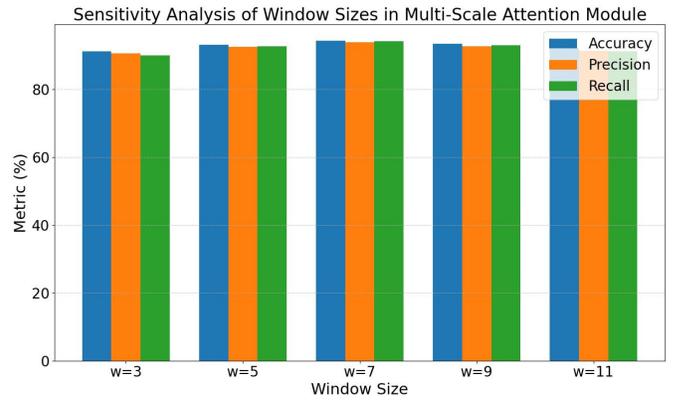

Figure 3. Sensitivity Analysis of Window Sizes in Multi-Scale Attention Module

As shown in Figure 3, the window size in the multi-scale attention module has a clear impact on model performance. When the window size increases from 3 to 7, the model shows consistent improvements in accuracy, precision, and recall. This suggests that larger windows can better capture temporal contextual information. They help the model identify more discriminative features and improve overall recognition performance.

When the window size is set to 7, all three metrics reach their highest values. This indicates a good balance between information coverage and local focus. A smaller window size (e.g., w = 3) may cause the model to capture only fragmented local features, limiting its ability to perceive global patterns. On the other hand, an overly large window (e.g., w = 11) may introduce redundant information. This can lead to dispersed attention and reduced focus, ultimately lowering the classification accuracy.

These experimental results further confirm the significant influence of the window parameter in the multi-scale mechanism. Proper window configuration can effectively enhance the model's ability to capture patterns at different granularities. Therefore, window size should be carefully tuned based on the characteristics of the data to achieve optimal mining performance and generalization ability.

## IV. Conclusion

This paper proposes a deep learning model that integrates Bidirectional LSTM with a multi-scale attention mechanism for sequence pattern mining tasks. The model aims to address the limitations of traditional methods in complex sequence modeling, such as weak feature extraction and incomplete context capture. Through empirical studies on multi-class sequence recognition tasks, the proposed model demonstrates superior accuracy and robustness. The results validate the effectiveness of combining bidirectional modeling with multi-scale perception for sequence information mining. The experiments evaluate the model's performance and stability from three aspects: model comparison, sensitivity to sequence length, and adjustment of attention window size. Results show that the proposed BiLSTM with multi-scale attention outperforms mainstream models across all performance metrics. It also adapts well to variations in input sequence length and attention window size, showing strong generalization ability and practical value.

In addition, the multi-scale attention module effectively guides the model to attend to key regions at different granularities. This prevents over-reliance on either local or global information and enhances the recognition of complex patterns. The framework offers a valuable reference for sequence analysis tasks in other domains, such as behavior recognition, text understanding, and multimodal fusion [19-21]. Future research may incorporate structure-adaptive mechanisms or graph attention modules to allow the model to dynamically adjust its perception strategy based on data characteristics [22]. Integration with contrastive learning or federated learning can also be explored. These directions may expand the model's potential in distributed environments, privacy-preserving applications, and cross-domain sequence mining, promoting the development of deep sequence modeling towards better generalization, interpretability, and robustness.